\documentclass[fleqn,11pt]{article} 
\usepackage{hyperref}
\usepackage{url}
\usepackage{amsmath,amsthm,amssymb}
\usepackage{graphicx}
\usepackage{algorithm}
\usepackage{algorithmic}

\usepackage{caption}
\usepackage{subcaption}

\usepackage{wrapfig}

\begin{document}

\title{Common Variable Learning and Invariant Representation Learning using Siamese Neural Networks}

\author{Uri Shaham \\  \href{mailto:uri.shaham@yale.edu}{uri.shaham@yale.edu} \\ Department of Statistics \\ Yale University 
\and Roy R. Lederman\thanks{The majority of this work was done while at the Applied Mathematics Program, Yale University.} \\ \href{mailto:roy@math.princeton.edu}{roy@math.princeton.edu} \\ The Program in Applied \& \\ Computational Mathematics \\ Princeton University}

\date{} 

\maketitle

\begin{abstract} 
We consider the statistical problem of learning common source of variability in data which are synchronously captured by multiple sensors, and demonstrate that Siamese neural networks can be naturally applied to this problem. 
This approach is useful in particular in exploratory, data-driven applications, where neither a model nor label information is available.
In recent years, many researchers have successfully applied Siamese neural networks to obtain an embedding of data which corresponds to a ``semantic similarity''.
We present an interpretation of this ``semantic similarity''  as learning of equivalence classes.
We discuss properties of the embedding obtained by Siamese networks and provide empirical results that demonstrate the ability of Siamese networks to learn common variability.

\end{abstract}


\section{Introduction}
\label{sec:intro}

Many machine learning and signal processing methods aim to separate desired variability (``signal'') from undesired variability (``noise'' and sensor idiosyncrasies). When the variability of interest is known on some set of data, supervised learning methods (such as regression) may be applied. Alternatively, when a data model is available, classic signal processing techniques (such as filtering) may be used to discard noise. 
The assumption that such knowledge is available is not always realistic; in many cases, in particular when exploring new data, it is not clear how to identify and represent the ``interesting part of the phenomenon''.
For example, in analysis of epileptic seizures we are interested in recovering patterns of activity that drive multiple areas of the brain, even though these patterns may be masked by massive nonlinear, non-additive effects of local activity, for which we have no model.

The purpose of this manuscript is to propose an approach for separating desired variability from irrelevant variability, in the absence of data model or label information. 
Our proposed approach is purely unsupervised, and is based on coincidence or co-occurrence, as a source of information from which the desired variability can be recovered. 
Specifically, we assume the data are measured through multiple sensors, and that the desired variability is recorded by both of the sensors, while the irrelevant effects are sensor-specific idiosyncrasies (i.e., each sensor records local phenomena and noise, independently of the other sensors).
We use coincidence (pairs that were measured at the same time by the different sensors) to recognize the common source of variability. 
The learning is performed using Siamese neural networks.

The modern form of Siamese neural networks have been proposed by \cite{chopra2005learning} to obtain an embedding of the input data that corresponds to ``semantic similarity'', so that Euclidean proximity between points in the embedding space implies that the points are ``semantically similar''. 
Siamese networks have been used for various tasks, such as dimensionality reduction, 
learning invariant representations~\cite{hadsell2006dimensionality}  and learning hashing functions~\cite{masci2014multimodal}.
This manuscript provides a formal model and a mathematical interpretation for the embedding that Siamese networks try to obtain, as representing equivalence classes.

Training of Siamese networks requires a collection of pairs of similar and dissimilar input objects;
we refer to the choice of these training collections as ``pairing''. 
In many works on Siamese networks, the pairing is based on information such as given classified samples, or on knowledge of a data model.
For example, In \cite{chopra2005learning} 
the pairing of objects is based on class membership, and the resulting representation is shown to be invariant to some input transformations such as pose and illumination of faces in images. 
In \cite{hadsell2006dimensionality} Siamese networks are applied to obtain a low dimensional embedding of the input data; the approach is based on computing a similarity graph of the input data; in the experiments, the graph is computed using Euclidean distance or knowledge of the model generating the data. We assume no such model. Furthermore, as discussed in Section \ref{sec:inputspace},
Euclidean proximity in the input space might not always capture the desired similarity.  
In \cite{masci2014multimodal} Siamese networks are used to obtain hash maps given a similarity measure; the experiments in the manuscript rely of class membership for pairing.
In \cite{taylor2011learning} Siamese networks are applied to obtain representations of images of people, which correspond to pose, and are invariant to undesired variability, such as identity and clothing; in this case the similarity calculation is based on having images where people were imitating the positions shown in a fixed set of seed images, i.e., on a known data model.
In \cite{yih2011learning} and \cite{hermann2014multilingual}, variants of Siamese architectures are applied to textual objects; in both cases the availability of pairs of objects labeled with a degree of similarity is assumed.

Our contributions are as follows: first, we use the formulation of the common variable learning problem \mbox{\cite{lederman2014common}} to interpret what Siamese network try to do. Specifically, we demonstrate that Siamese networks are in fact trained to recognize equivalence classes of an equivalence relation defined by the common variability we aim to learn. 
Put another way, the embedding obtained from a Siamese network corresponds to the quotient space of this relation. 
Second, we demonstrate how Siamese networks can  exploit coincidence as an efficient approach for separating desired variability from undesired variability in absence of neither data model nor label information.

The organization of this manuscript is as follows:
In Section~\ref{sec:learningByCo}, we  describe the common variable learning problem. 
In Section~\ref{sec:algorithm}, we show how coincidence can be used to train a Siamese network, and briefly review a typical training algorithm.
Some mathematical properties of the embedding which Siamese networks aim to obtain are discussed in Section~\ref{sec:discussion}.
In Section~\ref{section_results}, we present experiments using synthetic data, demonstrating that the common variable is indeed learned by Siamese networks.  
Brief conclusions are presented in Section~\ref{sec:conclusions}.


\section{Learning by Coincidence}\label{sec:learningByCo}

\subsection{Motivation}\label{sec:problem:motivation}
The purpose of this section is to illustrate the motivation for ``learning from coincidence''. We use a simplified toy example, adapted from \cite{lederman2014common}.
While this example appears to be a simple image processing problem, 
which is easily treated with some domain knowledge, 
we intentionally refrain from using this domain knowledge in order to demonstrate how the 
Siamese Networks work without the domain knowledge. 

\begin{figure}[ht!]
  \centering
    \includegraphics[height=3.5in]{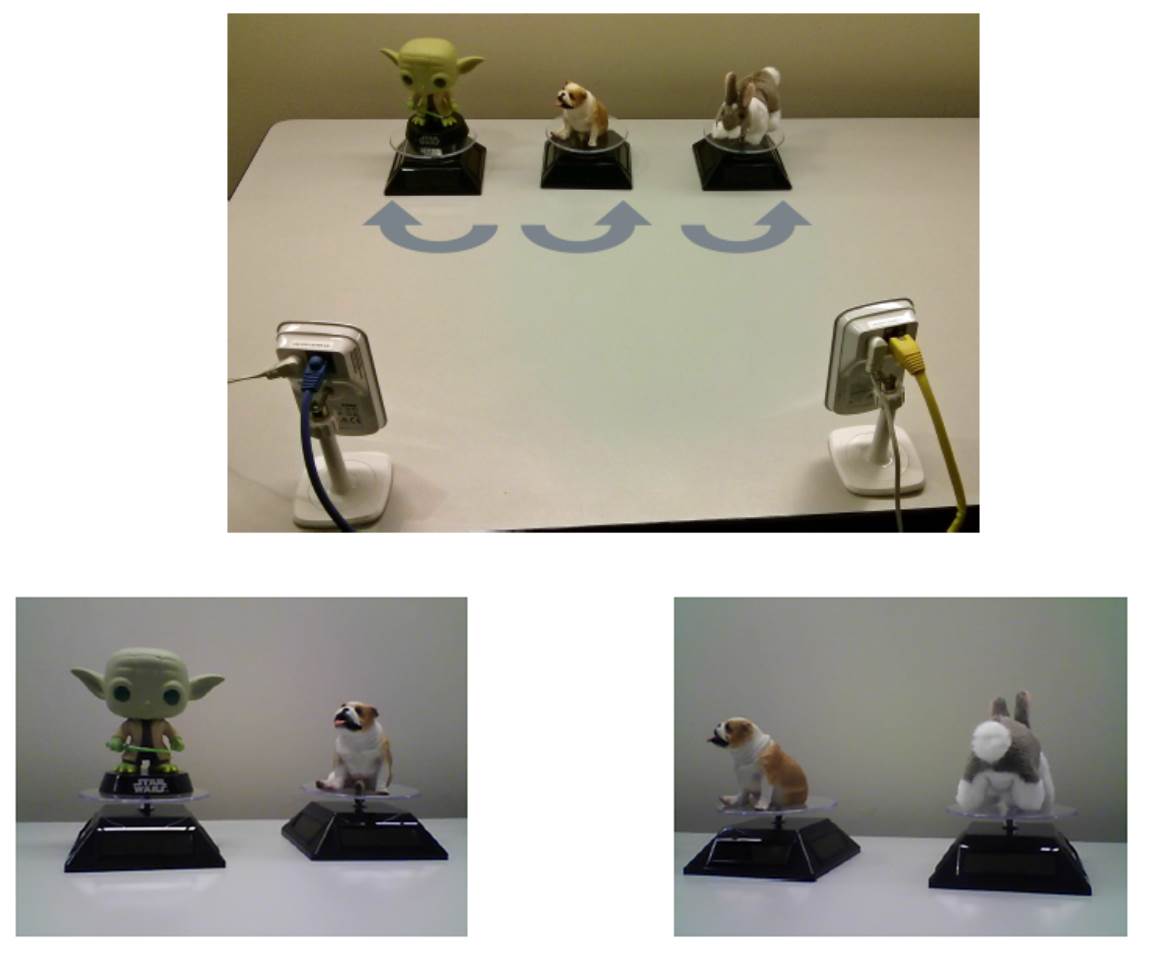}
   \caption{The Toy dataset experiment. Top: The experiment setting, in which Yoda, the bulldog and the bunny on spinning tables and the two cameras. Bottom: two sample snapshots, taken by cameras 1 (left) and 2 (right) at the same time. In both pictures the bulldog is in the same state $X$ (i.e., rotation angle, with respect to the table).
}
  \label{fig:bulldog}
\end{figure}

The experimental setup is presented in Figure \ref{fig:bulldog}.
Three objects, 
Yoda, a bulldog and a bunny, are placed on spinning tables and each object spins independently of the other objects.
Two cameras are used to take simultaneous snapshots: Camera 1, whose field of view includes Yoda and the bulldog, and Camera 2, whose field of view includes the bulldog and the bunny. In this setting, the rotation angle of the bulldog is a common hidden variable, which we will denote by $X$; the common variable is manifested in the snapshot taken by both cameras. 
The rotation angle of Yoda, which we will denote  by $Y$, is a sensor-specific source of variability manifested only in snapshots taken by Camera 1, 
and the rotation angle of the bunny, which we will denote by $Z$, is a sensor-specific source of variability manifested only in Camera 2. 
The three rotation angles are ``hidden'', in the sense that they are not measured directly, but only through the snapshots taken by the cameras.
Given snapshots from both cameras, our goal is to obtain a parametrization of the ``relevant'' common hidden variable $X$, i.e., the rotation angle of the bulldog, and ignore the ``superfluous'' sensor-specific idiosyncratic variables $Y$ and $Z$.

This specific task can be performed with specific knowledge of the problem, 
by masking the irrelevant objects, reducing the problem to learning of one variable. 
Indeed, when $X$ is the only variable influencing the measurements, 
there are various methods to explore the geometry of $X$ (e.g.  k-means, diffusion maps~\cite{coifman2006diffusion}). 
However, in the more interesting case, it is not known a-priori how the abstract value of $Y$ influences the measurements, and the measurements need not be images (for example, see Section~\ref{sec:diffMod}).
The scenario presented in this section is a simplified case of data-driven modeling and exploration of systems; $X$ represents some underlying phenomena which we would like to investigate although we don't have a model for the phenomena or for the other variables $Y$ and $Z$ that influence our measurements. 
In particular, we do not have any labeled examples of $X$.
Furthermore, we do not know in advance that it is possible to isolate $X$ by masking certain pixels (in fact, we do not know in advance that the samples represent images).  
The key to isolating $X$, from the superfluous $Y$ and $Z$ is the fact that we have multiple instances of measurements taken simultaneously by the two cameras;
in this case, while we don't know anything about the nature of the images, 
we know that $X$ (which turns out to be the bulldog) was in the same state when the two cameras took the snapshots.
More generally, such measurements that {\em have the same value of $X$}, give us a natural clue that we can use to reveal the structure of $X$. 

In this manuscript, we argue that the Siamese Networks, used in this context with the proper adaptations, attempt to recover a representation of $X$ that ignores the sensor specific $Y$ and $Z$.
We demonstrate that in this way, Siamese Networks can be used for unsupervised data-driven exploration of phenomena with very little domain knowledge, based mainly on examples that are obtained simultaneously, but without a prior model and without class labels or target values for regression.

\subsection{Common Variable Learning: Formal Definition}\label{sec:formal}
We follow a similar setting to \cite{lederman2014common}, which discussed the problem from a manifold learning perspective. 
Let $(X, Y, Z) \sim \pi_{x,y,z}(X, Y, Z)$ be three hidden random variables 
from the (possibly high dimensional) spaces $\mathcal{X}$, $\mathcal{Y}$ and $\mathcal{Z}$, 
with distribution $\pi_{x,y,z}$, where, given $X$, the variables $Y$ and $Z$ are independent. 

We have access to these hidden variables through two observable random variables 
$S^{(1)} = g_1(X, Y)$  and  $S^{(2)} = g_2(X, Z)$,
where $g_1$ and $g_2$ are bi-Lipschitz (therefore, invertible). We denote the range of $g_1$ and $g_2$ by $\mathcal{S}^{(1)}$ and $\mathcal{S}^{(2)}$, respectively; these ranges may be embedded in a high dimensional space.
We refer to the random variables $S^{(1)}$ and $S^{(2)}$ as the measurement in Sensor 1 and the measurement in Sensor 2, respectively. The $i$-th realization of the system consists of the hidden triplet $(x_i, y_i, z_i)$ and the corresponding pair of measurements  $(s^{(1)}_i,s^{(2)}_i )$;
while $x_i$, $y_i$ and $z_i$ are hidden and not available to us directly,  $s^{(1)}_i= g_1(x_i, y_i)$ and $s^{(2)}_i = g_2(x_i, z_i)$ are observable. 
We note that both $s^{(1)}_i$ and $s^{(2)}_i$ are functions of the same realization $x_i$ of $X$. Our dataset is composed of $n$ pairs of corresponding measurements $\left\{ (s^{(1)}_i, s^{(2)}_i ) \right\}_{i=1}^n$.

A natural way to obtain such pairs is to measure the same phenomenon $X$ with two different sensors $g_1$ and $g_2$, with both sensors influenced by the same phenomenon $X$, and each of them also influenced by its own idiosyncratic ``irrelevant'' state, $Y$ or $Z$.

Ideally, we would like to construct a function $\phi :\mathcal{S}^{(1)} \rightarrow \mathcal{X}$ that recovers $x$ from $g_1(x,y)$, so that for every $x\in \mathcal{X}$, and every $y\in \mathcal{Y}$, we would have $x=\phi \left( g_1(x,y) \right)$. 
However, since $\mathcal{X}$ and $g_1$ are unknown, we cannot expect to recover $x$ precisely, 
and we are interested in a function $f_1: \mathcal{S}^{(1)} \rightarrow \mathbb{R}^d$ that 
recovers $x$ up to some scaling and bi-Lipschitz transformation.
In particular, we require that for all $x \in \mathcal{X}$ and  $y,y' \in \mathcal{Y}$
\begin{equation}\label{equ:f1eq}
  f_1( g_1(x,y) ) =  f_1( g_1(x,y') ) ,
\end{equation} 
and for all $x \ne x' \in \mathcal{X}$ and  $y,y' \in \mathcal{Y}$
\begin{equation}\label{equ:f1neq}
  f_1( g_1(x,y) ) \ne  f_1( g_1(x',y') ) .
\end{equation}


\section{Algorithm}\label{sec:algorithm}

In this section we discuss the rationale for Siamese Networks in the context of the problem formulated above, and briefly review the Siamese Networks algorithm in this context. While the algorithm given in Section~\ref{sec:algo} is a typical variant of a Siamese network training algorithm, the key element of our approach is the implementation of learning through coincidence, which is manifested in the construction of the ``positive'' and ``negative'' datasets, described in Sections~\ref{sec:rational} and~\ref{sec:data}.


\subsection{Rationale}\label{sec:rational}

In order to satisfy Equations (\ref{equ:f1eq}) and (\ref{equ:f1neq}), we would like $f_1(s^{(1)}_i)$ to depend on $x_i$ and be invariant to the value of $y_i$. 
The crucial information is provided in the dataset through the fact that both $s^{(1)}_i$ and $s^{(2)}_i$
in the $i$-th pair are functions of the same value of $x_i$.
The idea is to use this information to learn maps $f_1$ and $f_2$ such that for all $i$,
\begin{equation}\label{equ:f1eqf2}
  f_1( s^{(1)}_i ) =  f_2( s^{(2)}_i ) .
\end{equation}
For every $x \in \mathcal{X}$, $y,y' \in \mathcal{Y}$ and  $z,z' \in \mathcal{Z}$, the functions $f_1$ and $f_2$ are required to satisfy
$f_1 (g_1 (x,y)) = f_1 (g_1 (x,y' ))$
and 
$f_2 (g_2 (x,z)) = f_2 (g_2 (x,z' ))$.
To avoid a trivial solution, in which $f_1$ and $f_2$ are simply constant functions, we add the requirement that for all $x_i \neq x_j$,
the functions also satisfy  
\begin{equation}\label{equ:f1neqf2}
  f_1( s^{(1)}_i ) \neq  f_2( s^{(2)}_j ),
\end{equation}
so that $f_1$ and $f_2$ cannot simply ``ignore'' the value of $x$.

We implement the function $f_1$ by a network which we denote by $\mathcal{N}_1$, 
and the function $f_2$ by a network which we denote by $\mathcal{N}_2$. 
 We are given a dataset $D_\text{pos}$ of ``positive'' pairs, $(s^{(1)}_i,s^{(2)}_i)$, in which both elements correspond to the same realization $x_i$ of the common variable $X$; in addition we are given (or construct) a dataset $D_\text{neg}$ of ``negative'' pairs,  $(\tilde{s}^{(1)}_i,\tilde{s}^{(2)}_i)$ in which the two elements correspond to different realizations of $X$ (see section \ref{sec:data}). 
 The idea is that when we introduce to the network a positive pair $(s^{(1)}_i,s^{(2)}_i) \in D_\text{pos}$
as input to $\mathcal{N}_1$ and $\mathcal{N}_2$, we require the outputs of $\mathcal{N}_1$ to be identical to the output of
$\mathcal{N}_2$,
whereas when we introduce to the network a negative pair $(\tilde{s}^{(1)}_i,\tilde{s}^{(2)}_i) \in D_\text{neg}$ 
as input, we require the output of $\mathcal{N}_1$ to be different from the output of $\mathcal{N}_2$.

At the end of the process, the map $f_1$, implemented by $\mathcal{N}_1$, computes our approximate representation;
as a useful ``side effect'', we also obtain the map $f_2$ which computes a similar approximate representation for the samples obtained from Sensor 2.

\begin{figure}[h!]
  \centering
    \includegraphics[height=2.8in]{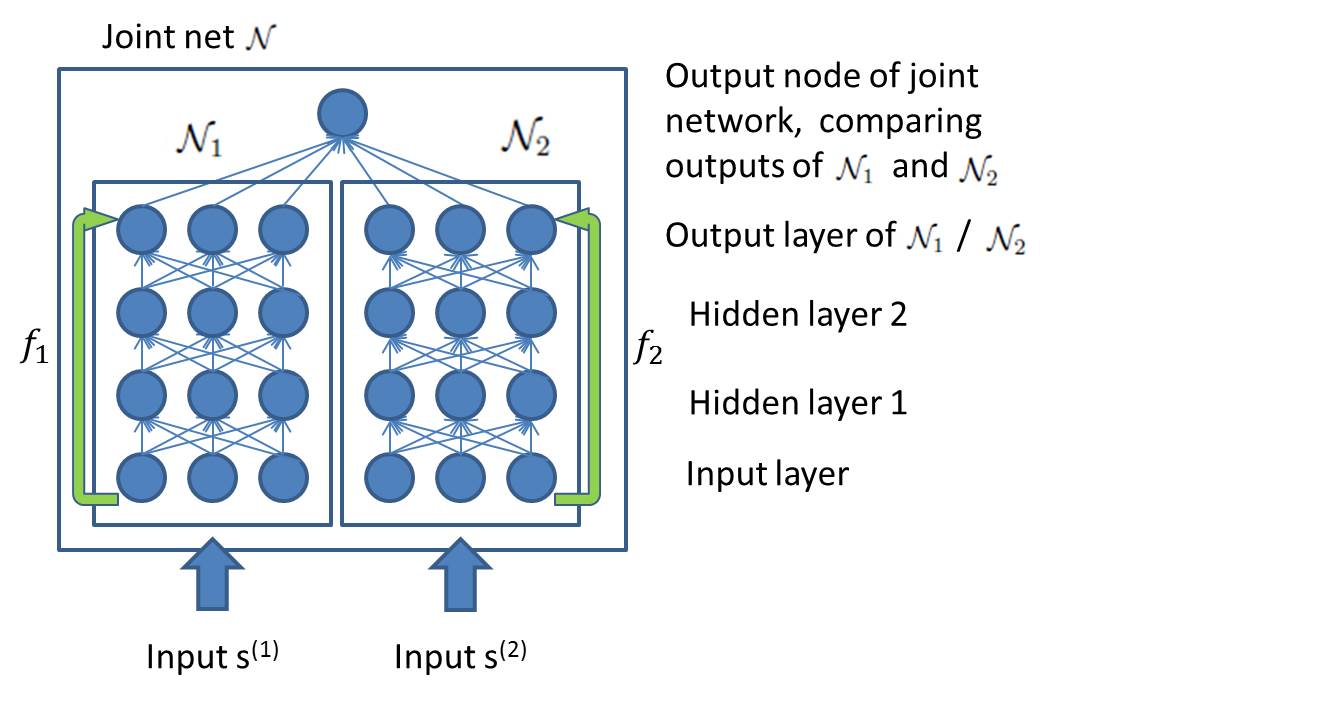}
   \caption{A diagram of the network structure.}
  \label{fig:bimodalNN}
\end{figure}

In summary, Siamese Networks try to achieve the following goal, when looking at pairs of measurements from two different sensors in the context of this manuscript:
\begin{itemize}
  \item If the two measurements have the same value of $X$ (measurements taken at the same time), give the same output.
  \item If the measurements probably don't have the same value of $X$, make the outputs of the two different networks different. 
\end{itemize}


\subsection{Implementing Coincidence: Constructing Datasets of Positive Pairs and Negative Pairs}
\label{sec:data}
The algorithm is given a dataset $D_\text{pos}$ of $n_\text{pos}$ pairs of the form $(s_i^{(1)},s_i^{(2)})$ corresponding to $n$ realizations of $X,Y,Z$. 
Hence, each instance $(s_i^{(1)},s_i^{(2)})\in D_\text{pos}$ is a pair of measurements that were taken at the same time by the two different sensors. We refer to this dataset as the {\em positive dataset}.
In addition, we construct a second dataset, referred to as the {\em negative dataset} $D_\text{neg}$, which contains $n_\text{neg}$ ``false pairs'', of the form $(\tilde{s}_{i}^{(1)},\tilde{s}_{i}^{(2)})$.
Ideally, $\tilde{s}_{i}^{(1)}$ and $\tilde{s}_{i}^{(2)}$ should 
be different realizations with different values of $X$, so that $\tilde{x}_{i} \ne \tilde{x}_{i}'$;
in practice, it suffices that $\tilde{x}_{i} \ne \tilde{x}_{i}$ with sufficiently high probability. When $D_\text{neg}$ is not explicitly available, an approximation is constructed 
from $D_\text{pos}$ by randomly mixing pairs from $D_\text{pos}$.

The training data for the Siamese network, (i.e., the positive and negative pairs) is obtained without assuming any class membership, data model or label information. The entire construction and pairing (as ``positive'' and ``negative'' pairs) is based on coincidence, i.e., on the fact that the data are measured through multiple sensors, with each of the sensors capturing the variability of interest.

\subsection{Algorithm}\label{sec:algo}

Algorithm~\ref{algo:1} is a typical training procedure for Siamese networks, presented here for completeness of the discussion and to provide the context for the discussion of pairing in Section~\ref{sec:data}.
 
\begin{algorithm}[h!]
   \caption{Common variable learning using an ANN}
   \label{alg:example}
\begin{algorithmic}
   \STATE {\bfseries Input:} $\{(s^{(1)}_i,s^{(2)}_i)\}_{i=1}^n $
   \STATE {\bfseries Output:} implementation of maps $f_1:\mathbb{R}^{d_1} \rightarrow \mathbb{R}^{d}$ and $f_2:\mathbb{R}^{d_2} \rightarrow \mathbb{R}^{d}$
   \STATE Construct datasets $D_\text{pos}$ and $D_\text{neg}$ (see Section \ref{sec:data})
   \STATE Optimize the parameters of the joint network $\mathcal{N}$ (equation \ref{eq:net:loss})
   \STATE Optional: dimensionality reduction of the learned representation 
\end{algorithmic}
\label{algo:1}
\end{algorithm}


A typical architecture of a Siamese network  $\mathcal{N}$ is presented in Figure \ref{fig:bimodalNN}.
The network $\mathcal{N}$ is composed of two networks $\mathcal{N}_1$ and $\mathcal{N}_2$ and a single output unit, which is connected to the output layer of both networks.  $\mathcal{N}_1$ and $\mathcal{N}_2$ accept samples from 
$\mathcal{S}^{(1)}$ and $\mathcal{S}^{(2)}$, respectively. The two networks may have different numbers of layers and different configurations; however, they have the same number $d$ of output units. 

The output node of $\mathcal{N}$ compares the output of $\mathcal{N}_1$ and $\mathcal{N}_2$ by computing 
$\sigma\left(\| f_1(s^{(1)}) - f_2(s^{(2)}) \|^2\right)$,
with $s^{(1)}$ and $s^{(2)}$ the inputs of $\mathcal{N}_1$ and $\mathcal{N}_2$, respectively,
$f_1(s^{(1)})$ and $f_2(s^{(2)})$ the outputs of $\mathcal{N}_1$ and $\mathcal{N}_2$, respectively,
and $\sigma$ the logistic sigmoid function $\sigma(x) = \frac{1}{1 + e^{-x}}$.

In our experiments in Section~\ref{section_results} we set the training loss function to be
\begin{equation}\label{eq:net:loss}
\begin{split}
   &L(\theta) = \\
   &\frac{\alpha}{n_\text{pos}}  \sum_{(s^{(1)},s^{(2)})\in D_\text{pos}}   \left( \frac{1}{2}- \sigma\left(\| f_1(s^{(1)}) - f_2(s^{(2)}) \|^2\right) \right)^2  + \\
   &\frac{\beta}{n_\text{neg}}  \sum_{(s^{(1)},s^{(2)})\in D_\text{neg}} \left(1-\sigma\left(\| f_1(s^{(1)}) - f_2(s^{(2)}) \|^2\right) \right)^2  + \\
   &\lambda \|\theta\|_2^2, 
   \end{split}
\end{equation}
 where $\theta$ is a vector containing the weight parameters (but not bias parameters) of $\mathcal{N}_1$ and $\mathcal{N}_2$. 
For the positive pairs in $D_\text{pos}$, we would like $\| f_1(s^{(1)}) - f_2(s^{(2)}) \|^2$ to be close to zero, 
thus $\sigma \left( \| f_1(s^{(1)}) - f_2(s^{(2)}) \|^2 \right)$ close to $\sigma(0)=\frac{1}{2}$;
similarly, for the negative pairs in $D_\text{neg}$, we would like to maximize $\| f_1(s^{(1)}) - f_2(s^{(2)}) \|^2$, 
thus have $\sigma \left( \| f_1(s^{(1)}) - f_2(s^{(2)}) \|^2 \right)$ close to $\sigma(\infty)=1$.

Once the network $\mathcal{N}$ is  trained, $\mathcal{N}_1$ and $\mathcal{N}_2$ implement our proposed functions $f_1$ and $f_2$, respectively.

The network $\mathcal{N}$ bears some superficial resemblance to a classifier that determines whether or not two 
measurements from two different sensors share the same value of $X$ (i.e. ``real'' pairs or ``fake'' pairs).  
However, classifiers need not construct a representation of the common variable, which is the goal in this work. 
In addition, since we do not use any class membership information, and the entire training is based on coincidence, our proposed training approach is purely unsupervised.
Having said that, in our experiments we find it useful to measure the ``classification accuracy'' of the network as a proxy for the quality of learning.
Specifically, since the output of the network ranges between $0.5$ and $1$, we set $0.75$ as a classification threshold for estimating whether $(s^{(1)}, s^{(2)})$ is a ``real'' or a ``fake'' pair.


\section{Discussion}\label{sec:discussion}

\subsection {Siamese Networks Learn Equivalence Classes}
\label{sec:quotient}

Let $\sim$ be an equivalence relation on the  $\mathcal{S}^{(1)}$, the space of measurements in Sensor 1. We say that two observations are equivalent if and only if they share the same value of $X$, i.e., $s^{(1)}_i \sim s^{(1)}_j \text{ iff } x_i=x_j$.
This equivalence relation generates the quotient set $\mathcal{S}^{(1)} / \sim$, where the equivalence class of $s^{(1)} = g_1(x,y)$ is $[s^{(1)}] = \left\{  g_1(x,y'): y' \in \mathcal{Y} \right\}$.

We observe that a function $f_1$ that satisfies (\ref{equ:f1eq}) yields the same value 
for any member of an equivalence class $[s^{(1)}]$.
Moreover, a function $f_1$ that satisfies (\ref{equ:f1neq}) also yields a different value for members of
different equivalence classes  $[s^{(1)}_i] \ne [s^{(1)}_j] \in \mathcal{S}^{(1)} / \sim$. 

Thus, with a minor abuse of notation, there is a natural way to define $f_1 : \mathcal{S}^{(1)} / \sim \rightarrow \mathbb{R}^d$ on the quotient set $\mathcal{S}^{(1)} / \sim$ rather than on $\mathcal{S}^{(1)}$.
Furthermore, such $f_1$ is an injective function. Hence, 
ideally, the function $f_1$ implemented by (a single sub-net of a) Siamese network is effectively a map of the equivalence class of its argument. 

\subsection {Comparing Measurements from the Same Sensor}\label{sec:topology_f1}

In practice, because of the continuity of the functions $g_1$ and $g_2$ and the continuity of the 
computation operations in the networks that we use here,  
samples that are ``close'' in $X$ would have similar representations, 
so that the representation of $X$ is smooth. Informally, 
\begin{equation}
x_i \approx x_j \Leftrightarrow f_1(g_1(x_i,y_i)) \approx f_1(g_1(x_j,y_j)),
\end{equation}
therefore, the function $f_1$ can be used to estimate if two samples $s^{(1)}_i$ and $s^{(1)}_j$ in Sensor 1 have ``close'' values $x_i$ and $x_j$.

Therefore, from this perspective, the vague ``semantic similarity'' is interpreted as proximity in the space $\mathcal{X}$ of the common variable.

\subsection {Measurements from Different Sensors Become Comparable}\label{sec:compf1f2} \label{sec:compaccuracy}

The algorithm treats the measurements in Sensor 1 and the measurements in Sensor 2 symmetrically,
in the sense that it aims to construct maps $f_1 : \mathcal{S}^{(1)} \rightarrow \mathbb{R}^d$ and 
$f_2 : \mathcal{S}^{(2)} \rightarrow \mathbb{R}^d$ that map into the same codomain $\mathbb{R}^d$ 
and have similar properties. 
Moreover, the algorithm aims to find such $f_1$ and $f_2$ that agree in the sense defined in equations (\ref{equ:f1eqf2}) and (\ref{equ:f1neqf2}).

Following the same argument as in Section \ref{sec:topology_f1}, the two functions $f_1$ and $f_2$ 
can be used to compare a sample $s^{(1)}_i=g_1(x_i,y_i)$ from Sensor 1 to a sample $s^{(2)}_j=g_2(x_j,z_j)$ from Sensor 2 to estimate whether the two samples are obtained from ``close'' values of $X$; informally,
\begin{equation}\label{eq:compare_test}
x_i \approx x_j \Leftrightarrow f_1(g_1(x_i,y_i)) \approx f_2(g_2(x_j,z_j)) .
\end{equation}
The two sensors might measure different modalities, such as audio signals in one and images in the other, 
so that the framework proposed here allows to compare two different modalities in terms of the common variable.

Several works propose ANNs that learn representations of inputs that are measured via two sources, possibly of different modalities, for example audio and video, or images and texts \cite{ngiam2011multimodal, srivastava2012multimodal}. These works focus on learning a shared representation, containing information from the two modalities, and demonstrate that one modality provides information about the other modality. These works have been particularly interested in recovering the input in one modality from the input in another modality, for example, through an autoencoder.
The problem of learning a shared representation of objects which are captured via multiple sensors, possibly of different modalities is discussed also in~\cite{keller2010audio}, where diffusion maps are used to obtain the representation. However, here as well sensor specific variability is not removed. 
In this manuscript, we aim to discard modality-specific attributes, and learn the common hidden variable that underlies both modalities.

\subsection{Similarities in the Input Space}\label{sec:inputspace}

One of the interesting properties of the common variable problem, demonstrated in the toy example in Figure \ref{fig:bulldog}, and in the examples in the next section, is that similarity in the common variable, or ``semantic similarity'', can have very little to do with the similarity in the input space. 
For example, in the toy problem, we can have two different snapshots taken by Camera 2 which are supposed to be equivalent because the bulldog (the common variable $X$) happens to be in the same place.
However, the bunny, which is actually a larger, more dominant object, may appear in any state in the two snapshots, so the snapshots are very different in the input space. 
In other words, snapshots that are very different in the input space may be equivalent. 
Similarly, snapshots in which the bunny appears in a similar state but the bulldog does not will be similar in the input space, although they are not equivalent in the sense of the common variable we wish to capture. 
Therefore, similarity in the input space (measured via, say, Euclidean distance) might have little to do with similarity in the common variable, and consequently is an inappropriate tool for collecting ``positive'' pairs in this context. 

Some Siamese networks use the distance in the input space to define pairs (e.g. nearest neighbors in the input space are paired in \cite{hadsell2006dimensionality}), or to regularize the distance in the output space; this use of similarity in the input space has been demonstrated to be useful in dimensionality reduction.  
However, this type of pairing or regularization cannot discard the superfluous variables because it cannot distinguish between the superfluous variables and the common ones. Therefore, pairing based on the simultaneous measurements, when such measurements are available, is advantageous in recovering the common variable and in distilling hidden underlying phenomena. 


\subsection{Connection to CCA}\label{sec:dcca}
A natural approach for discovering common information in a dataset of paired observations is to use Canonical Correlation Analysis (CCA) \cite{hotelling1936relations}. The ability of standard CCA to discover such information is limited, since it only considers linear transformations of its inputs. Among non linear versions of CCA, the Deep-CCA architecture proposed in \cite{andrew2013deep} bares resemblance to Siamese Network architecture.
In this manuscript we follow a different approach in the use of the dataset, 
and an architecture that resembles Siamese Networks more than Deep-CCA.
Our experiment in section~\ref{sec:dcca} suggests that the approach proposed in this manuscript better suits the common variable learning problem.


\subsection{Learning Invariant Representations using Siamese Networks}\label{sec:invariant_problem}

In this section we will discuss a related problem, learning invariant representation, 
which can also be viewed as a problem of learning equivalence classes. 

Let $G$ be a group that acts on a set $\mathcal{S}$. 
We say that $s \in \mathcal{S}$ is equivalent to $s' \in \mathcal{S}$ up to $G$ if there is $g \in G$ such that $g.s = s'$ .
We denote the equivalence relation by $s \sim s'$.
We say that a map $f$ of $\mathcal{S}$  is invariant to $G$ if it satisfies (a) for all $g \in G$, 
  $f( s ) =  f( g.s )$
and (b) for all $s \not \sim s'$ and for all $g \in G$,  $f( s ) \ne  f( g.s' )$.

In the invariant representation learning problem, we have examples of pairs $( g.s , g'.s)$ with different randomly selected group actions $g,g' \in G$ operating on a randomly selected element $s \in \mathcal{S}$
and in some cases we may have examples of ``negative pairs'' $( g.s , g'.s' )$ with $s \not \sim s' \in \mathcal{S}$; 
we would like to use such examples to find a function $f$ that is invariant to $G$. In other words, we would like to learn a map that is defined on the equivalence classes of $s \sim s'$. 
Since Siamese networks learn equivalence classes, they can be used to learn invariant representation, as demonstrated in Section \ref{sec:CTrot}.

Neural networks which are invariant to specific input transformations, such as translation and rotation have been proposed in \cite{ranzato2007unsupervised}, \cite{oyallon2014deep},  and \cite{sohn2012learning}), for example. 
However, these networks are often designed to be invariant to specific, well modeled transformations, rather than to unknown transformations.


\section{Experimental Results}
\label{section_results}

In this section we present experimental results of common variable learning. The experiments involve synthetic datasets, generated so that the common variable is defined explicitly, to demonstrate that the embedding obtained by the net indeed corresponds to the quotient space defined by the common variable. We also demonstrate how Siamese networks can be used to learn invariant representations.

In experiments where we have more than a single hidden layer in each stack, we pre-train every hidden layer in $ \mathcal{N}_1$ and  $ \mathcal{N}_2$ as a Denoising Autoencoder (DAE)\cite{vincent2008extracting} with activation sparsity loss (see \cite{UFLDL1}).
The optimization of the network $\mathcal{N}$ is performed using standard Stochastic Gradient Descent (SGD) with momentum (see, for example, \cite {sutskever2013importance}) and dropout (see, for example, \cite{dahl2013improving}), or using L-BFGS (see, for example, \cite{wright1999numerical}); 
in both optimization algorithms, we compute gradients using standard backpropagation (see, for example, \cite{rojas1996neural}). 
The classification accuracy we report is measured on a test set consisting of positive and negative examples that were not introduced to the network during training. 


\subsection{Common Variable Learning: the Toy Dataset (Spinning Figures)}\label{sec:exp:spin_ii}

\begin{figure}[ht!]
  \centering
     \includegraphics[height=2.5in]{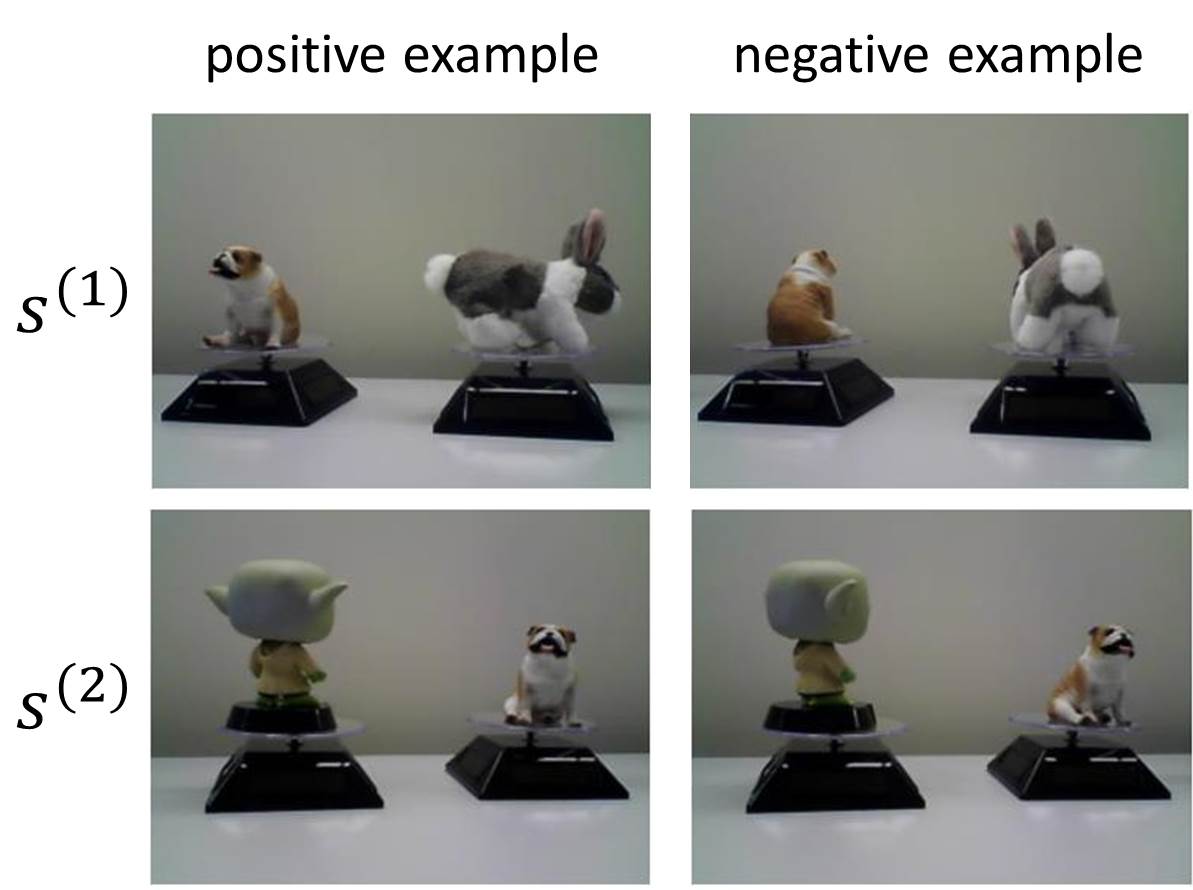}
     \includegraphics[height=2.5in]{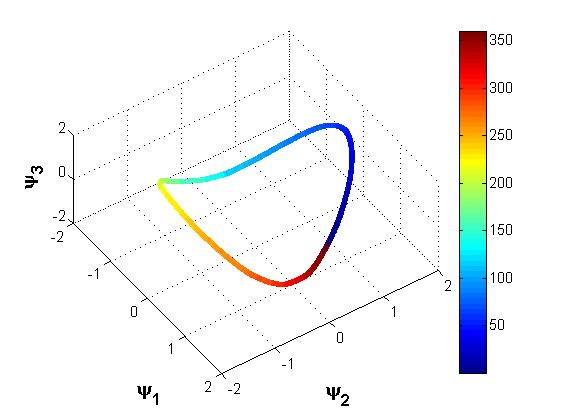}
   \caption{The Toy dataset experiment.
   Top: two sample examples. In a positive example snapshots taken simultaneously, containing two different views on the bulldog at the same rotation angle.  In a negative example the snapshots from the two cameras were not taken at the same time, hence they do not correspond to same rotation angle of the bulldog. 
   Bottom: embedding of the Toy dataset. The color of each point corresponds to the true common hidden variable, i.e. the rotation angle $X$ of the bulldog.
}
  \label{fig:bulldogEmbedding}
\end{figure}

In this experiment we revisit the setup described in Section \ref{sec:problem:motivation}. 
Here, $s^{(1)}$ is a snapshot taken by Camera 1 and $s^{(2)}$ is a snapshot taken by Camera 2.
The dataset $D_\text{pos}$ consists of pairs of snapshots $(s^{(1)}_i , s^{(2)}_i)$, with $s^{(1)}_i$ and $s^{(2)}_i$ taken 
simultaneously by Camera 1 and Camera 2, respectively. 
The dataset $D_\text{neg}$ was constructed 
by pairing snapshots that had been taken at different times. 
The samples $s^{(1)}_i$ and $s^{(2)}_i$ are $60 \times 80$ color images; 
positive and negative examples are presented in Figure \ref{fig:bulldogEmbedding} (top).
The training sets $D_\text{pos}$ and $D_\text{neg}$ consisted of $10,000$ examples each.
Both $\mathcal{N}_1$ and  $\mathcal{N}_2$ had three layers,
the two hidden layers in each network had $150$ units, 
and the output layers had $100$ units. 
The joint network $\mathcal{N}$ was trained using L-BFGS. 
The classification accuracy on the test set (as defined in Section \ref{sec:algo}) was  95.96\%. 

The learned representations in this experiment (the outputs of the networks $\mathcal{N}_1$ and $\mathcal{N}_2$), are $100$-dimensional. We used standard dimensionality reduction algorithms to process the output for the purpose of visualization and further processing; in Figure {\ref{fig:bulldogEmbedding}} (bottom) we present the reduced representation obtained using diffusion maps \cite{coifman2006diffusion}, which we found to be clearer than the representation obtained using PCA.
The closed curve and the smooth transitions in color in the embedding demonstrate that the algorithm recovered a good representation of the common variable $X$, and that the position along the learned manifold corresponds to the value of the common variable. 

\subsection{Common Variable Learning: Two Different Modalities}\label{sec:diffMod}

In the previous experiment, we used the same type of input in both sensors; 
in this experiment we used a different data modality in each sensor: 
images in one sensor and audio signals in the other. 

\begin{figure}[ht!]
  \centering
    \includegraphics[height=2.0in]{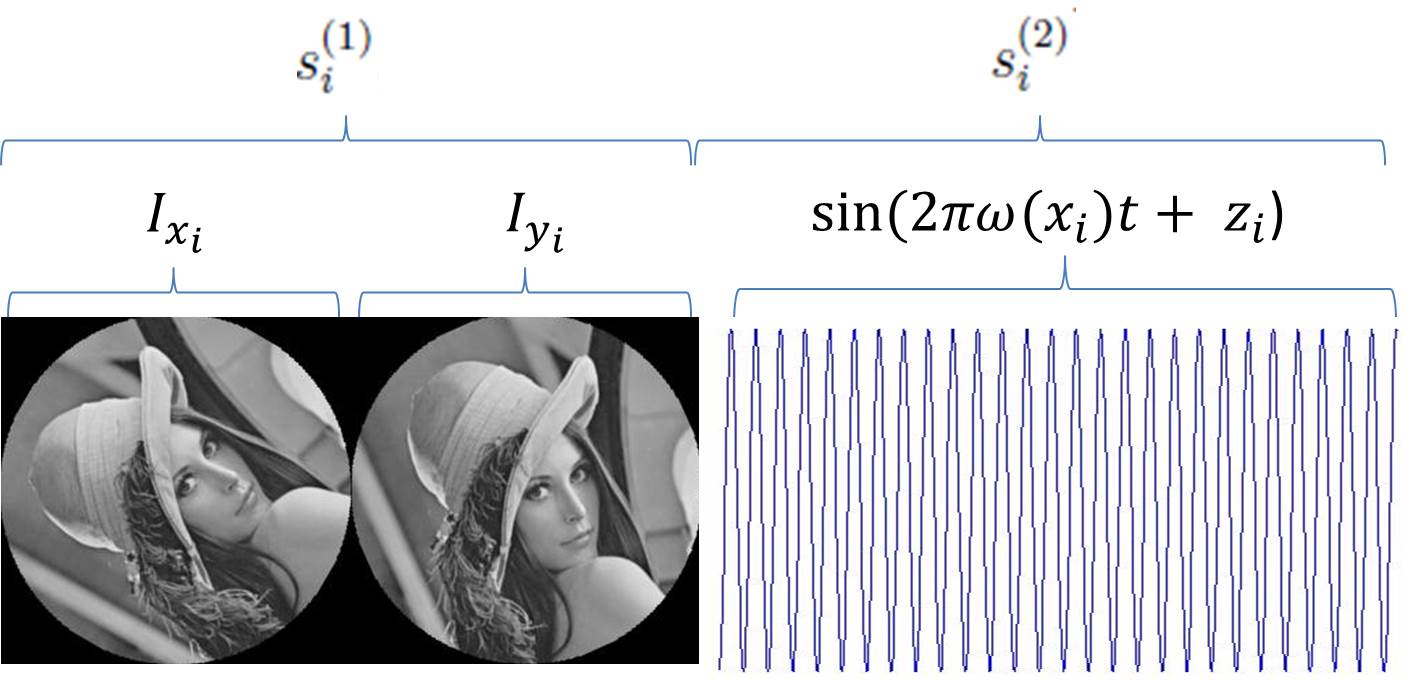}
    \includegraphics[height=2.5in]{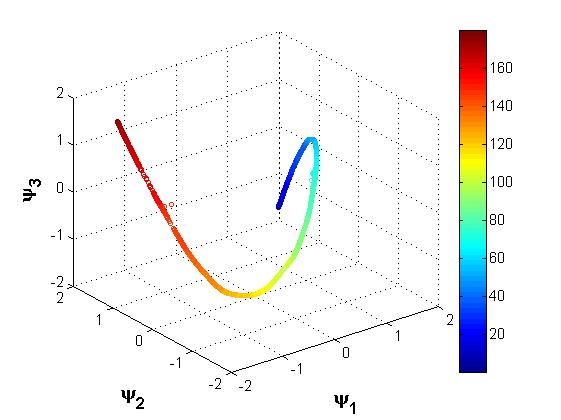}
   \caption{The two modalities dataset. Top: a positive example from the dataset. A measurement $s^{(1)}_i$ from sensor 1 is a concatenation of two rotations of the image $I$, in angles $x_i$ and $y_i$. A measurement $s^{(2)}_i$ from sensor 2 is a sine, with frequency determined by $x_i$ and phase determined by $z_i$. Bottom: embedding of the images and audio signals from the test set in the two modalities experiment; the color corresponds to the true value of the common variable $x_i$. Data from two different modalities is mapped to the same space, and is parametrized by the common variable. }
  \label{fig:mmEmbedding}
\end{figure}

We denote by $I_\theta$ an image $I$ rotated by angle $\theta$. A measurement $s^{(1)}_i$ from sensor 1 is a concatenation $(I_{x_i},I_{y_i})$ of two rotations of an image in arbitrary angles . A measurement $s^{(2)}_i$ from sensor 2 is a $T$ dimensional vector 
$v_{x_i,z_i}(t),\; t=1,..,T$
with entries
$v_{x_i,z_i}(t) = \sin (2 \pi \omega(x_i)t + z_i)$,
where $\omega(\cdot)$ is a deterministic function, so that $x_i$ determines the frequency of the sine, 
and $z_i$ determines the phase. 
In other words, the common variable $X$ determines the rotation of  the left image in the first sensor and the frequency of the sine in the second sensor;
the sensor specific variables are the rotation angle of the right image in $s^{(1)}_i$, and the phase of the sine in $s^{(2)}_i$. An example from the dataset of this experiment is presented in Figure \ref{fig:mmEmbedding} (top).

Both $\mathcal{N}_1$ and $\mathcal{N}_2$ had three layers, with $100$ units in each. 
$D_\text{pos}$ and $D_\text{neg}$ consisted of $10,000$ examples each. 
The accuracy on the test set was $95.8\%$. 

In Figure \ref{fig:mmEmbedding} (bottom) we present the diffusion embedding of the outputs of both $\mathcal{N}_1$ and $\mathcal{N}_2$, colored by the true value of the common variable $x_i$;
the smooth transition of the color along the manifold implies that the learned representation corresponds to the common variable.
Furthermore, the points in Figure \ref{fig:mmEmbedding} (bottom) which correspond to output of $\mathcal{N}_1$ are
indistinguishable from the points that correspond to outputs of  $\mathcal{N}_2$; in other words, data from the two different modalities has been mapped into a the same space, where data points from same or different modalities can be compared based on their corresponding value of the common variable $x_i$.

\subsection{Learning a Rotation-Invariant Representation}
\label{sec:CTrot}

The following experiment demonstrates the application of the algorithm to the problem of learning invariant representations, discussed in Section~\ref{sec:invariant_problem}.

\begin{figure}[h!]
  \centering
    \includegraphics[height=2.3in]{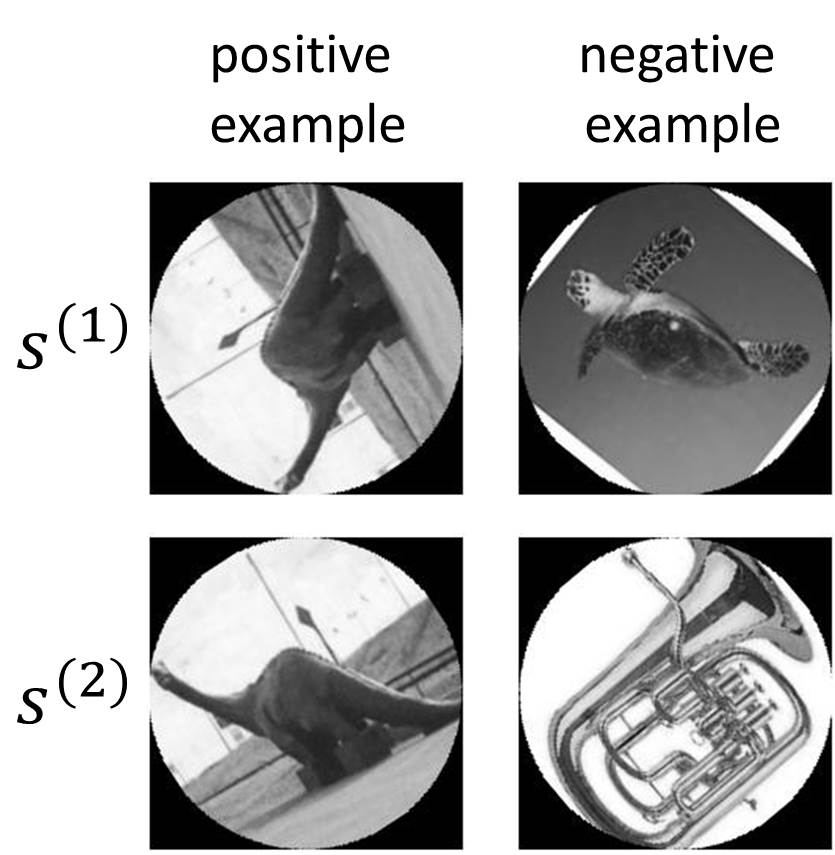}
    \includegraphics[height=2.3in]{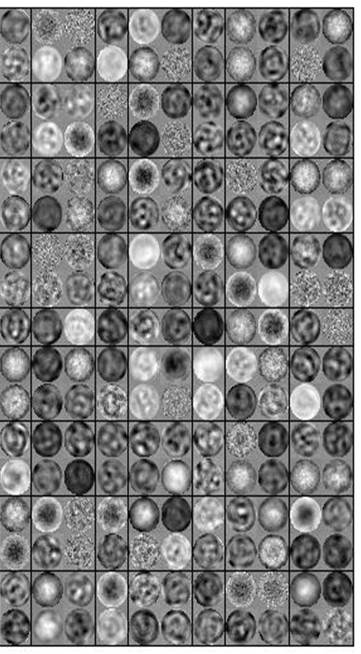}
    \includegraphics[width=2.7in]{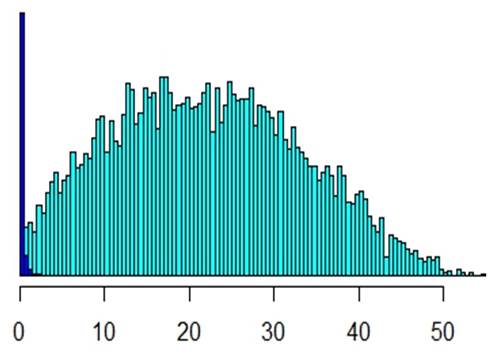}
   \caption{The rotation-invariance experiment. Top left: two sample examples generated from the Caltech-101 dataset for the rotation-invariance experiment. In a positive example $s^{(1)}$ and $s^{(2)}$ are the same image, up to rotation.  In a negative example $s^{(1)}$ and $s^{(2)}$ are the different images, rotated in different angles. Top right: first layer features in the rotation-invariance experiment. Bottom: Histograms of  $\|f_1(a)-f_1(a')\|_2$ (dark blue) and $\|f_1(a)-f_1(b)\|_2$ (light blue). Distances between representations of arbitrarily rotated copies of the same image are significantly smaller than distances between representations of different images.} \label{fig:CaltechFeatures}  \label{fig:Caltech}
\end{figure}

Our goal here is to learn maps $f_1$ and $f_2$ that are rotation-invariant.
Let $G$ be the group of rotations of images, so that $g.s$ is a rotation of $s$ by $g$ degrees. 
We used images from the Caltech-101 dataset \cite{fei2007learning} (converted to gray-scale $50 \times 50$ pixels for convenience), and constructed datasets of rotated images. 
The positive set $D_\text{pos}$ was composed of samples $(s^{(1)}_i,s^{(2)}_i)$ where $ s^{(1)}_i=g_{1,i}.s_i$ and $ s^{(2)}_i=g_{2,i}.s_i$ are two instances of the same base image $s_i$ rotated by two randomly chosen angles $g_{1,i},g_{2,i} \in G$.
Each sample in the negative dataset was composed of two different randomly chosen base images, each rotated by a different randomly chosen angle. Positive and negative examples from the dataset and the first layer weights of $\mathcal{N}_1$ are presented in Figure {\ref{fig:CaltechFeatures}} (top).

The networks $\mathcal{N}_1$ and $\mathcal{N}_2$ had three  layers each; the joint network was trained using L-BFGS. The learned functions achieved a high accuracy score of 99.44\%; 

To check whether the hidden representation we obtained is indeed invariant to rotation, we performed the following analysis: we randomly selected an image and rotated it in two random  angles; we denote the resulting images by $a$ and $a'$. 
We then selected a different image and rotated it in a random angle; we denote the resulting image by $b$. 
If the map $f_1$ is indeed invariant to rotations, then we expect to have
$\|f_1(a)-f_1(a')\|_2  \ll \|f_1(a)-f_1(b)\|_2$. 
Histograms of $\|f_1(a)-f_1(a')\|_2$ and $\|f_1(a)-f_1(b)\|_2$ for 10,000 repetitions of the above procedure are presented in Figure \ref{fig:Caltech} (bottom);
as evident from the histograms, $\|f_1(a)-f_1(a')\|_2$ is indeed significantly smaller than $\|f_1(a)-f_1(b)\|_2$, as expected.

\clearpage

\subsection{Comparison to Deep CCA}\label{sec:dcca}
\begin{figure}[h!]
  \centering
  \begin{center}
        \includegraphics[height=2.7in]{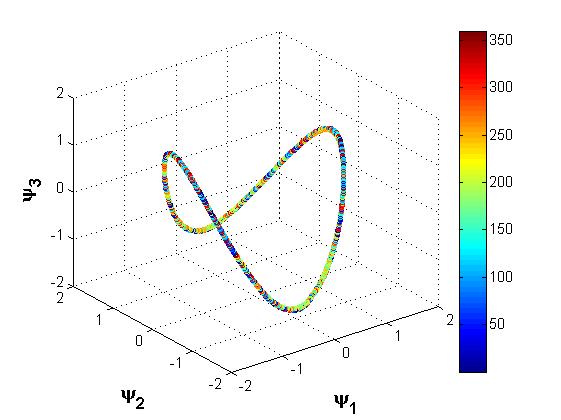}
  \end{center}
  \caption{Diffusion embedding obtained from the deep CCA network \cite{andrew2013deep} on the Toy dataset. 
The color of each point corresponds to the value of the common hidden variable,
an embedding that captures the common variable would have a smooth transition of color;
the embedding here does not correspond to the value of the common hidden variable.}
  \label{fig:dcca}
\end{figure}

Given realizations $\{(s^{(1)}_i,s^{(2)}_i)\}_{i=1}^n$ of random variables $S^{(1)}$ and $S^{(2)}$, the deep CCA algorithm (see \cite{andrew2013deep}) computes maps $f'_1$ and $f'_2$ so that the cross correlation between $f'_1(S^{(1)})$ and $f'_2(S^{(2)})$ is maximized. 
We implemented the deep CCA network 
and applied it to the Toy dataset of Section \ref{sec:exp:spin_ii}, with the same network structure used in our experiment in Section \ref{sec:exp:spin_ii}.
The diffusion embedding that was obtained from the last layer representation of the deep CCA network is presented in Figure \ref{fig:dcca}. We observe that in this experiment the position along the embedded manifold does not correspond to the value of the common variable, i.e., the rotation angle of the bulldog; moreover, additional analysis indicated that the representation obtained by deep CCA in this experiment reflects the sensor specific superfluous variables (rotation angles of Yoda and the bunny), which we would like to discard. 


\section{Conclusions}
\label{sec:conclusions}

In this manuscript we presented Siamese neural networks as a solution to the statistical problem of common variable learning. 
We demonstrated that 
Siamese neural networks learn equivalence relations in the input space. 
We demonstrated how coincidence can be used for the recovery of common variables, in the absence of a model or labeled data,
using examples of measurements that are ``equivalent'' or ``related'' via an appropriate form of coincidence, and using examples of measurements that are ``not equivalent'' or ``unrelated''.
In addition, we demonstrated how a Siamese network can map observations, possibly from different modalities, to a space in which their respective values of the common variable are comparable.

The experiments presented in this manuscript have been carefully designed to illustrate the theoretical arguments regarding the embedding obtained by Siamese networks representing the common variable and regarding limited use of domain knowledge.
As demonstrated in other works, when domain knowledge is available it can be used in designing the network architecture: for example, when the samples are images, it is natural to use convolutional networks.


\section*{Acknowledgments}

The authors would like to thank Raphy Coifman,  Sahand N. Negahban,  Andrew R. Barron
and Ronen Talmon, for their help.

\bibliography{common_v10}
\bibliographystyle{ieeetr}

\end{document}